%% file: Main.tex
\documentclass[sigconf]{acmart}

\pdfoutput=1
\usepackage{soul}
\usepackage{enumitem}
\usepackage{amsmath}

\usepackage{hyperref}
\usepackage{graphicx}

\copyrightyear{2023}
\acmYear{2023}
\setcopyright{rightsretained}
\acmConference[CIKM '23]{Proceedings of the 32nd ACM International Conference on Information and Knowledge Management}{October 21--25, 2023}{Birmingham, United Kingdom}
\acmBooktitle{Proceedings of the 32nd ACM International Conference on Information and Knowledge Management (CIKM '23), October 21--25, 2023, Birmingham, United Kingdom}
\acmDOI{10.1145/3583780.3614824}
\acmISBN{979-8-4007-0124-5/23/10}

\makeatletter
\gdef\@copyrightpermission{
  \begin{minipage}{0.3\columnwidth}
 \href{https://creativecommons.org/licenses/by/4.0/}{\includegraphics[width=0.90\textwidth]{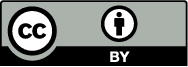}}
  \end{minipage}
  \hfill
  \begin{minipage}{0.7\columnwidth}
\href{https://creativecommons.org/licenses/by/4.0/}{This work is licensed under a Creative Commons Attribution International 4.0 License.}
  \end{minipage}
  \vspace{5pt}
}
\makeatother

\begin{document}

\title{Contrastive Learning of Temporal Distinctiveness for Survival~Analysis~in~Electronic Health Records}

\author{Mohsen Nayebi Kerdabadi}
\email{mohsen.nayebi@ku.edu}
\affiliation{%
  \institution{University of Kansas}
  \city{Lawrence}
  \state{KS}
  \country{USA}
}

\author{Arya Hadizadeh Moghaddam}
\email{a.hadizadehm@ku.edu}
\affiliation{%
  \institution{University of Kansas}
  \city{Lawrence}
  \state{KS}
  \country{USA}
}

\author{Bin Liu}
\email{bin.liu1@mail.wvu.edu}
\affiliation{%
  \institution{West Virginia University}
  \city{Morgantown}
  \state{WV}
  \country{USA}
}

\author{Mei Liu}
\email{mei.liu@ufl.edu}
\affiliation{%
  \institution{University of Florida}
  \city{Gainesville}
  \state{FL}
  \country{USA}
}

\author{Zijun Yao}
\email{zyao@ku.edu}
\authornote{Corresponding author.}
\affiliation{%
  \institution{University of Kansas}
  \city{Lawrence}
  \state{KS}
  \country{USA}
}

\renewcommand{\shortauthors}{Mohsen Nayebi Kerdabadi, Arya Hadizadeh Moghaddam, Bin Liu, Mei Liu, \& Zijun Yao}

\begin{abstract}
Survival analysis plays a crucial role in many healthcare decisions, where the risk prediction for the events of interest can support an informative outlook for a patient's medical journey. Given the existence of data censoring, an effective way of survival analysis is to enforce the pairwise temporal concordance between censored and observed data, aiming to utilize the time interval before censoring as partially observed time-to-event labels for supervised learning. Although existing studies mostly employed ranking methods to pursue an ordering objective, contrastive methods which learn a discriminative embedding by having data contrast against each other, have not been explored thoroughly for survival analysis. Therefore, in this paper, we propose a novel Ontology-aware Temporality-based Contrastive Survival (OTCSurv) analysis framework that utilizes survival durations from both censored and observed data to define temporal distinctiveness and construct negative sample pairs with varying hardness for contrastive learning. Specifically, we first use an ontological encoder and a sequential self-attention encoder to represent the longitudinal EHR data with rich contexts. Second, we design a temporal contrastive loss to capture varying survival durations in a supervised setting through a hardness-aware negative sampling mechanism. Last, we incorporate the contrastive task into the time-to-event predictive task with multiple loss components. We conduct extensive experiments using a large EHR dataset to forecast the risk of hospitalized patients who are in danger of developing acute kidney injury (AKI), a critical and urgent medical condition. The effectiveness and explainability of the proposed model are validated through comprehensive quantitative and qualitative studies.
\end{abstract}



\maketitle

\newcommand{\model}{OTCSurv}
\newcommand{\clloss}{SupWCon}

\input{tex/introduction}
\input{tex/related_work}

\input{tex/methodology}

\input{tex/experiments}
\input{tex/interpretability}

\input{tex/Conclusion}

\bibliographystyle{ACM-Reference-Format}
\balance
\bibliography{references}

\end{document}

%% file: tex/introduction.tex
\section{Introduction}
The increasingly abundant electronic health records (EHRs) have provided an unprecedented opportunity to apply predictive analytics to support healthcare decisions \cite{yadav2018mining,rajkomar2018scalable}. To achieve the optimal outcomes for a patient's medical journey, an important question faced by healthcare providers is how to precisely anticipate the adversarial events (e.g., kidney injury, heart failure, and stroke), so that these critical incidents can be responded to timely with sufficient clinical attention. 
Therefore, it is crucial to investigate the application of survival analysis (SA) in longitudinal EHR data, which aims to identify the significant factors that influence the degree of risks and to further forecast the time to events of interest.

For survival analysis, a key challenge is how to deal with the existence of censored data for time-to-event modeling. In the case of censoring, events of interest may not be observed for some patients due to the limited duration of observation or the withdrawal of patients during the study.
In order to address this challenge, various traditional survival analysis models have been developed although they suffer from multiple limitations. Parametric survival models \cite{mittal2013large} assume a specific distribution for the baseline hazard function, such as the exponential, Weibull, or log-normal distribution. However, events in the real world are usually too complex to be captured by such predefined distributions.
On the other hand, although the semi-parametric Cox model \cite{cox1972regression} makes no assumptions about the baseline hazard function, it requires the hazard function to be multiplicatively proportional to the covariates. Moreover, most of these approaches (e.g., Cox-based models) only focus on predicting the relative ordering of survival durations of individuals, overlooking their actual event time. Therefore, the capability of time estimation for future event occurrences is unfortunately compromised.

To overcome the limitations of early studies, deep learning techniques have been increasingly applied to survival predictive tasks \cite{ren2019deep,giunchiglia2018rnn, lee2018deephit} which offer the capacity to capture complex survival patterns without making explicit distributional assumptions. While some studies have explored the enforcement of patient concordance by survival probabilities to accommodate both observed and censored survival data, there exists very limited literature on contrastive learning (CL) methods aimed at learning a discriminative representation of patient records to achieve better predictive performance.
In contrastive learning, data are contrasted against each other in self-supervised \cite{chen2020simple}, semi-supervised \cite{zhang2022semi}, or supervised \cite{khosla2020supervised} settings. Generally, it trains an objective to distinguish the subtle characteristics in data, by maximizing the similarity between positive pairs (instances that belong to the same labels) and minimizing the similarity between negative pairs (instances that have different labels). Although the positive vs. negative labeling strategy for contrasted pairs has been defined by self-augmentation (i.e., whether the pair originates from a single data point) or supervised classes \cite{hong2022deep} (i.e., whether the pair belongs to the same class), the exploration of contrasting labeling based on temporal distinctiveness (which is based on the time difference between the two survival durations) for survival analysis is still lacking. Furthermore, given that survival duration is a numerical entity, accounting for the hardness defined by the time difference for contrastive labels can help the model learn the survival data with more flexibility.

Another challenge associated with survival analysis in EHR is the possible data insufficiency. Usually, a large variety of medical codes are recorded in a dataset, but many codes may have a relatively small number of occurrences (e.g., rare diseases). As a result, for patients with rare codes or sparse visits, the embedding of their medical history is often sub-optimal. One way to address this issue is to incorporate the domain-specific knowledge inherent in medical ontology into the representation of EHR features \cite{yao2023ontology}. Medical ontology is a hierarchical classification structure of medical concepts (e.g., diagnosis, medications, etc.), which can serve as an auxiliary categorization for knowledge representation \cite{choi2017gram, ma2018kame, song2019medical}. For example, GRAM \cite{choi2017gram} proposes a graph-based attention model that employs the attention mechanism on hierarchical levels of each medical code to learn medically meaningful EHR feature embeddings. With ontological encoding, survival models can better build the association between codes or patients, and transfer the medical knowledge from one sample to another. Therefore, to further improve the quality of patient profiling, the ontology learning of EHR features can be integrated with the contrastive learning core of survival analysis.

In this paper, we introduce an Ontology-aware Temporality-based Contrastive Survival analysis framework called \model{}, which combines the ontology-enhanced EHR data encoder, the contrastive learning of temporal distinctiveness, and the survival probability predictor with multiple loss components, for interpretable, data-efficient, and discriminative survival analysis.
Specifically, the main contributions of this study can be summarized in three-fold:
\begin{itemize}
    \item We design a Supervised Weighted Contrastive (\clloss{}) Learning loss function that uses survival duration as its pairing criteria which is able to utilize both observed and censored observations. \clloss{} considers the hardness of negative pairs based on the survival duration differences to enrich the grain of contrastive learning.
    \item We used a sequential attention-based ontological encoder to learn medically informed embeddings for sequential hospital visits of patients. Ontology information brings data efficiency to our model by referring to higher-level medical concepts when the observation is sparse. 
    \item We optimize survival prediction through multiple loss components, focusing on two key goals: accurately predicting survival duration time and precisely ranking the risks or survival probabilities of patients at each time point. We train our model with a meticulous configuration of \clloss{}, accompanied by three more loss functions, guiding the training towards an optimum point satisfying these two goals.
\end{itemize}

Finally, we evaluate our proposed method and demonstrate the strength of our model on a real-world EHR dataset for Acute Kidney Injury (AKI) by performing baseline comparison, ablation study, and interpretability analysis.  

%% file: tex/related_work.tex
\section{Related Work}
In this section, we provide an overview of key contributions and advancements in the survival analysis field, concentrating on relevant methodologies and techniques in the literature.

One of the most widely used statistical methods in survival analysis is the Kaplan-Meier (KM) estimator \cite{kaplan1958nonparametric} which is a non-parametric survival analysis method, calculating the survival probability by dividing the number of individuals who have survived up to a given time by the number of patients at risk just before that time. However, KM does not take into account the covariates of patients. Early works in survival analysis primarily revolved around the Cox proportional hazards (CPH) model \cite{10.2307/2985181}, which assumes a proportional relationship between covariates and the hazard function. Due to the advantages of CPH, such as simplicity and interpretability, many survival analysis models have been proposed based on CPH, such as incorporation of time-varying covariates \cite{lin2002modeling}, accounting for competing risks \cite{fine1999proportional}, and CoxTime \cite{JMLR:v20:18-424} which expands upon Cox model by extending its capabilities beyond the assumption of proportional hazards.

In recent years, there has been an increasing interest in applying machine learning techniques to survival analysis. Random Survival Forests \cite{ishwaran2008random}, Deep Exponential Families \cite{ranganath2016deep, ranganath2015deep}, and semi-parametric Bayesian models based on Gaussian Processes \cite{fernandez2016gaussian}, offer flexibility in capturing complex survival patterns and handling non-linear relationships. As for deep learning-based  approaches, DeepSurv \cite{katzman2018deepsurv} introduced the application of deep neural networks for survival prediction, capturing complex relationships between covariates and survival outcomes using the Cox partial likelihood loss function. This has opened up many doors for utilizing deep learning in survival analysis, leading to the development of models like DeepHit \cite{lee2018deephit}, which is a multitask deep learning model capable of handling competing risks, DRSA \cite{ren2019deep} and RNN-SURV \cite{giunchiglia2018rnn}, both of which exploiting a recurrent neural network (RNN) to handle sequential data, and N-MTLR \cite{fotso2018deep} which leverages deep neural networks to replace the linear core of the MTLR \cite{yu2011learning}.

Some more recent state-of-the-art deep learning-based SA models are Dynamic-DeepHit \cite{lee2019dynamic}, Survtrace \cite{wang2022survtrace}, and Deep-CSA \cite{hong2022deep}. An extension of DeepHit is Dynamic-DeepHit which instead of the simple neural network, uses a recurrent neural network to dynamically capture longitudinal dependencies in the presence of competing risks. Survtrace proposes a transformer-based SA model that handles competing risks and benefits from a multi-task learning framework to learn a strong shared representation. Transformer-Based Deep Survival Analysis \cite{hu2021transformer} tries to make a trade-off between time predictive power and risk ranking power using both the absolute error as well as ranking evaluation metrics.

Generally, the existing architectures suffer from multiple limitations. Some works, such as Cox-based survival models, show suboptimal performances due to certain assumptions for the underlying stochastic process. Violations of these assumptions can lead to incorrect conclusions. Some of the deep learning methods are black boxes and do not offer sufficient interpretability. Also, many works only employ ranking methods to reach survival rate concordance and, to the best of our knowledge, there is no exploration of the use of contrastive methods based on the temporality for healthcare survival analysis. OTCSurv managed to mitigate the aforementioned challenges and support interpretable, data-
efficient, and discriminative survival analysis. As demonstrated in the following sections, OTCSurv exhibits an enhanced performance compared to its predecessors.

%% file: tex/methodology.tex
\section{Proposed Method}
In this section, we first describe the notations and formulate the EHR survival analysis problem. We then present the overview of the model. Last, we introduce each module in detail.

\begin{figure*}[th]
\begin{center}
\includegraphics[width=1.0\linewidth]{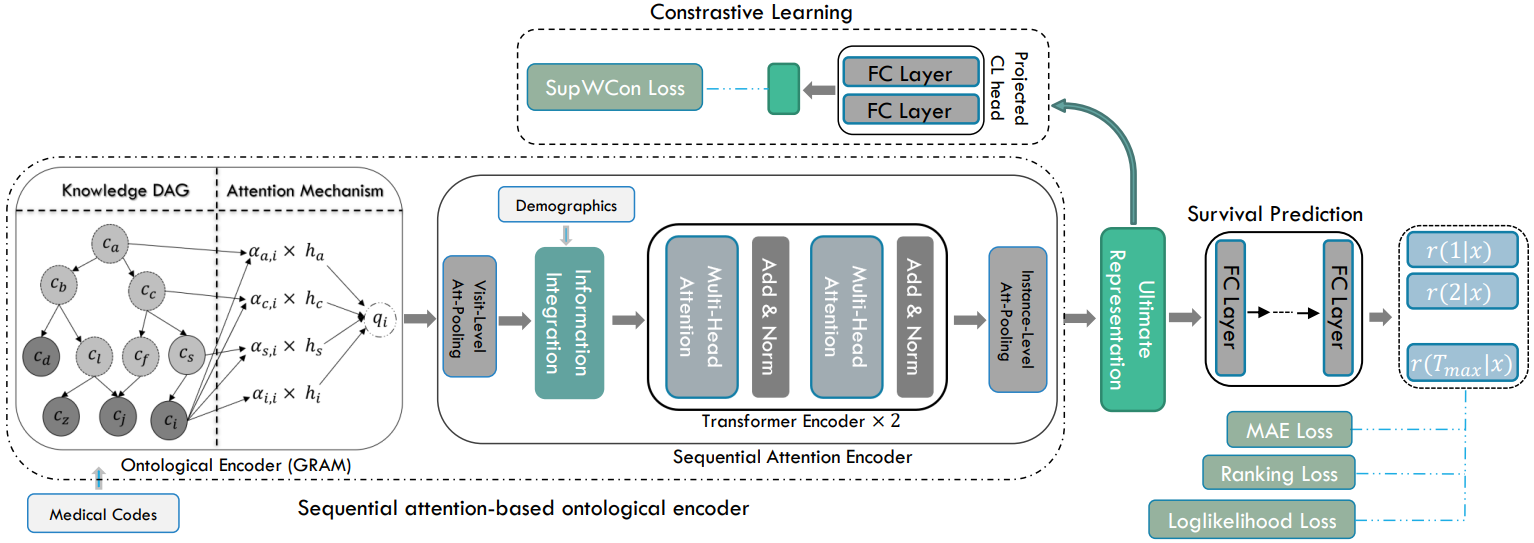}
\vspace{0.1cm}
\captionsetup{skip=1pt}
\caption{Architecture of the proposed \model{} model. There are three main components: 1) Sequential attention-based ontological encoder, which mainly consists of an ontological encoder, two attention-pooling blocks, and a transformer encoder to learn the ultimate instance-level representations of patients. 2) Contrastive Learning, which uses an intermediary transformation to transfer the ultimate representations to another latent space where SupWCon is functioning. 3) Survival prediction, which is a fully connected neural network and outputs the probabilities necessary for survival analysis calculation.}
\label{fig:Model}
\end{center}
\end{figure*}

\subsection{Problem Formulation}
Electronic healthcare records (EHRs) usually contain comprehensive information about a patient's medical history. Each patient normally has multiple hospital visits where diagnoses, prescriptions, and procedures are recorded using standardized codes in the hospital's database. EHRs can be exploited in three sets of information to be used in survival analysis: 1) covariates, 2) time to the event, and 3) a label indicating the type of the event (censored/observed). A discrete and finite time window with a maximum length of $T_{\text{max}}$ is considered for the time prediction. Therefore, our goal is to predict in which time interval $t \in \{0, \ldots, T_{\text{max}}\}$ the event of interest is most likely to happen, or to determine the probability of survival in each time interval. We show the event label by a binary variable $k$. If the instance is observed $k = 1$, otherwise (censored) $k = 0$. We can consider each instance (i.e., patient) as a triple of $(V,t,k)$ where $V =\{v_{n}\}_{n=1}^{N}$ is a sequence of covariates showing $N$ visits, and $v_{n} =  \{c_{1},c_{2}, \ldots,c_{|C|}, d_{1},d_{2}, \ldots,d_{|\mathcal{D}|}\}$ indicating the existence of both binary and continuous features. Binary medical codes are denoted by $c_{i}$, and continuous features such as demographics are denoted by $d_{j}$ where $|C|$ and $|\mathcal{D}|$ represent the sizes. For each medical code $c_{i}$, we extract the set of its ancestor codes (higher level concepts) in the hierarchy of the medical ontology, represented as a directed acyclic graph (DAG).

We denote the probability by $P$, the hazard function by $\lambda(t)$,  the probability density function by $f(t)$, and the survival probability by $S(t)$. By adding the caret symbol to each notation, we indicate their estimated forms, e.g., $\hat{S}(t)$ is the estimated survival probability. 

\textbf{Task:} Given the patient's sequential medical history in terms of longitudinal hospital visits containing medical codes, we aim to build a model to estimate the survival probability of patients in each time interval inside the prediction time window in the future. 

\subsection{Model Overview}
In this subsection, we introduce an overview of our proposed \model{} model architecture. As shown in Figure \ref{fig:Model}, the model consists of three main components. The first component is \textbf{Sequential attention-based ontological encoder} that consists of two main blocks. The first one is the ontological encoder which effectively utilizes the inherent valuable information within the medical ontologies to generate informed embedding vectors for medical codes. Next is the sequential attention encoder which consists of three attention-based parts: visit-level attention-pooling, transformer encoder, and instance-level attention-pooling. The visit-level attention-pooling uses the attention mechanism to reduce the dimension of the visit representations. Then, the output of the visit-level attention-pooling integrates with demographic information inside a data integration block to produce a representation containing all the patient's information. This representation along with positional encoding of visits is fed to the transformer encoder. The multi-head attention of the transformer will extract the interactions of medical visits to produce a rich representation of a patient that encompasses all the meaningful information. Instance-level attention-pooling is implemented on the output of the transformer encoder, to compress and combine the information of different visits of an instance using the attention mechanism and produce the ultimate instance representation. This ultimate instance representation will go through the second and the third main components of \model{} parallelly. One is \textbf{Contrastive Learning} component (the second main component) where a projection head which is a nonlinear transformation, e.g., a simple multilayer perceptron (MLP) with a nonlinear activation function, transfers the ultimate instance representation to a different latent space. This is where our proposed \clloss{} loss comes into play, adding temporal distinctive refinements to the ultimate representations. In parallel, the ultimate representation is fed into the \textbf{Survival Prediction} component (the third main component) which consists of a fully connected neural network. This neural network predicts $T_{\text{max}}$ number of probabilities, which are the complement of the hazard rates, for each of the $T_{\text{max}}$ predefined time intervals. To train this model, combined with SupWCon, three loss functions of \textbf{Loglikelihood Loss}, \textbf{Pairwise Ranking Loss}, \textbf{Mean Squared Error loss} are implemented to guide the model towards an optimum point regarding predictive, discriminating, and ranking ability in survival analysis.

\subsection{Sequential Attention-based Ontological Encoder}
The sequential attention-based ontological encoder is responsible for generating instance representations using attention-based components which are explained hereunder. 

\subsubsection{Ontological Encoder}
In order to address the challenge of data limitation in the healthcare domain, acquire comprehensive representations of medical codes, and increase predictability, we utilize the attention-based graph representation approach known as GRAM \cite{choi2017gram}. First, an initial embedding vector $h_{j} \in \mathbb{R}^{d_{c}}$ is assigned to each medical code as well as its ancestors (higher lever concepts) in the medical ontology, where $d_{c}$ is the code embedding dimension. Then, each code's final representation $q_{i} \in \mathbb{R}^{d_{c}}$ is calculated as a convex combination of the initial embeddings of itself and its ancestors using the attention mechanism:  
\begin{equation}
q_{i} = \sum_{j \in A(i)} \alpha_{ij}h_{j},\ \   \sum_{j \in A(i)} \alpha_{ij} = 1,\ \ \alpha_{ij} \geq 0  \ \ for \ \ j \in A(i)
\end{equation}
where $A(i)$ is the set containing the indices of the code $c_{i}$ and its ancestors. $\alpha_{ij} \in \mathbb{R}^{+}$ shows the attention weight given to ancestor code embedding $c_{j}$ when calculating $q_{i}$, which is the final representation of $c_{i}$. Using a Softmax function, $\alpha_{ij}$  is formulated as:

\begin{equation}
\alpha_{ij} = \frac{\exp(f(h_{i}, h_{j}))}{\sum_{k \in A(i)} \exp(f(h_{i}, h_{k})}
\end{equation}
\begin{equation}
f(h_{i}, h_{k}) = \omega^{T}_{\alpha}\tanh({W_{\alpha}Concat(h_{i};h_{k}) + b_{\alpha}})
\end{equation}
where $Concat(h_{i};h_{k})$ is the concatenation of $h_{i}$ and $h_{j}$ in a child-ancestor order. $f(\cdot)$ is an MLP operator with learnable parameters of $\omega_{\alpha}, W_{\alpha}, b_{\alpha}$.

\subsubsection{Attention-Pooling}
We used two attention-pooling components in our architecture, one after the ontological encoder, which is the visit-level attention-pooling, and one after the transformer encoder, which is the instance-level attention-pooling, to compress the information flow using the attention mechanism.

Assumes that the input of visit-level attention-pooling for a patient $i$ is a tensor $E_{i} \in \mathbb{R}^{N \times M \times d_{c}}$, where $N$, $M$, $d_{c}$ are the number of visits, the specified maximum number of possible codes inside each visit, and the dimension for the code embedding\footnote{We omit the patient index $i$ in the following notation for easier demonstration.}, respectively. Using the attention mechanism, we assign a weight to each code in a visit and use those weights to calculate the weighted average of medical code vectors. Thus instead of having a vector of size $(N \times M \times d_{c})$ for each patient, we reduce its dimension to a vector of size $(N \times d_{c})$. So, given the $n$-th visit representation $v_{n} \in \mathbb{R}^{M\times d_{c}}$, which is the concatenation of M code embeddings $q^{n}_{m}$ ($1\leq m \leq M$), we calculate an attention energy $e^{n}_{m} \in \mathbb{R}$ for each of $M$ medical code embedding:
\vspace{-2pt}
\begin{equation}
   e^{n} = l(v_{n}) = W_{2}\sigma(v_{n}W_{1} + \beta_{1}), \ \ 1 \leq n \leq N
\end{equation}
where $e^{n} \in \mathbb{R}^{M\times 1}$ contains M attention energies for the codes within $n$-th visit, and $l(\cdot)$ is a MLP operater with a ReLU activation function $\sigma$ and learnable parameters of $W_{1}$, $W_{2}$, $\beta_{1}$. Using softmax on attention energies, we calculate attention weights $\alpha^{n} \in \mathbb{R}^{M \times 1}$:

\vspace{-5pt}
\begin{equation}
    \alpha^{n} = \text{softmax}(e^{n})
\end{equation}
where $\alpha^{n}$ is the concatenation of M attention weights $\alpha_{m}^{n}$ ($1\leq m \leq M$). Finally, we have 

\vspace{-7pt}
\begin{equation}
    p^{n} = \sum_{i=1}^{M} \alpha^{n}_{m} q_{m}^{n}
\end{equation}
where $p^{n} \in \mathbb{R}^{d_{c}} (1 \leq n \leq N)$ represents the $n$-th visit of the patient. So, for each patient, we have $P \in \mathbb{R}^{N\times d_{c}}$ as the concatenation of $N$ visit representations $p^{n} \in \mathbb{R}^{d_{c}}$. The output of the visit-level attention block $P \in \mathbb{R}^{N\times d_{c}}$ is concatenated with each patient demographic embedding $s \in \mathbb{R}^{N\times d_{s}}$ ($d_{s}$ is the dimension for the demographic feature embedding) to obtain $F = \text{Concat}(P, s) \in \mathbb{R}^{N\times d}$ where $d = d_{c}+d_{s}$.

For the instance-level attention-pooling which is implemented on the output of the transformer encoder, we use the same technique described above to reduce the dimensionality. The output of the transformer encoder for a patient is $U \in \mathbb{R}^{N \times D}$, where $D$ is the transformer dimension. $U$ is fed to the instance-level attention-pooling, where using the attention mechanism, $N$ attention weights for each of the visit representations are generated. These weights are used to calculate the weighted average of visit representations, thereby reducing the dimension of $U$, outputting $u \in \mathbb{R}^{D}$ as the ultimate instance (patient) representation to be used in both the contrastive task and the survival prediction downstream task.  

\subsubsection{Transformer Encoder}
The encoder of the transformer architecture serves as the primary block for obtaining representations for survival analysis. For each patient, the input to the encoder of the transformer is a sequence of final visits' embeddings. The transformer's multihead-attention mechanism captures complex relationships among different hospital visits of a patient, enabling the model to encode comprehensive information about their dependencies over time. This results in rich representations that can capture survival patterns and time-dependent features.

\subsection{Temporal Distinctiveness with Supervised Weighted Contrastive Learning}
\label{sec:cl}
Contrastive learning aims to learn meaningful representations by maximizing agreement between similar examples while minimizing agreement between dissimilar examples. One technique to measure the similarity between two vectors is the cosine similarity, which can be calculated by the dot product of two vectors. It calculates the cosine of the angle between two vectors, representing their similarity by assessing how closely the two vectors align in the vector space. In this study, we formulate a contrastive learning loss function featuring an adaptive temperature parameter, referred to as Supervised Weighted Contrastive (SupWCon) loss. SupWCon is an extended version of the method proposed in \cite{khosla2020supervised} and has been tailored for survival analysis, particularly for handling censored data. We formulate
\vspace{-5pt}
\begin{equation}
    L^{\text {SupWCon }}=\sum_{i \in I} \frac{-1}{|P(i)|} \sum_{p \in P(i)} \log \frac{\exp \left(\boldsymbol{z}_i \cdot \boldsymbol{z}_p / \tau\right)}{\sum_{a \in A(i)} \exp \left(\boldsymbol{z}_i \cdot \boldsymbol{z}_a / \tau_{ia}\right)}
\end{equation}
where $I$ is the set of indices of all the instances, $P(i)$ is the set of indices of the instances that make a positive pair with the instance $i$, and $A(i) \equiv I$ \textbackslash $\{i\}$. The dot ($\cdot$) operator in the formulation represents the dot product of two vectors. $\tau \in \mathbb{R}^{+}$ is the constant scalar temperature parameter for positive pairs and $\tau_{ia} \in \mathbb{R}^{+}$ is the adaptive scalar temperature parameter for negative pairs which will be explained shortly. The instance with the index $i$ is called the anchor. Positive and negative pairs were particularly generated considering both the survival duration times and the labels of instances (observed/censored). For anchor $i$, which is an observed instance, any other observed instance with the survival duration time $t$ (which is the duration from day one to the day before the event time) inside the time window of $T_{i} - T/2 \leq t < T_{i} + T/2$ (referred to as positive window) makes a positive pair with the anchor and belongs to $P(i)$. Time window length $T$ is a hyperparameter that needs to be tuned with respect to the $T_{\text{max}}$, data distribution, and the nature of the problem. Any observed instance with a survival duration time $t$ outside the positive window ($t < T_{i} -T/2$ or $T_{i} +T/2 \leq t$) plus any censored instance with the survival duration time (which is the duration from day one to the day of censoring) greater or equal to $T_{i} + T/2$ ($T_{i} + T/2\leq t$) makes a negative pair with the anchor. We do not consider censored instances with a censoring time smaller than $T_{i} + T/2$ for both positive and negative pair generation because what happened to the patient after censoring is unknown (whether they were diagnosed with AKI or not, and if so when that happened). In fact, if, after censoring, the event (AKI) happens inside the positive window of the anchor, making a negative pair is wrong. Conversely, for patients with a censoring time greater or equal to $T_{i} + T/2$, we are sure that their survival duration is outside of the positive window of the anchor, so they are safe to be considered for negative pair generation. 

The temperature parameter for positive pairs $\tau$ is a constant positive scalar for all positive pairs and will be chosen by hyperparameter tuning. However, we adjusted the temperature parameter for each negative pair to encourage our model to regulate the amount of dissimilarity between the representations of negative pairs that exhibit various differences in survival duration. Hence, the model can better capture the distinction for negative pairs of various hardness. For example, if patient $i$ and patient $j$ make a negative pair, the adjusted temperature parameter for this negative pair is calculated as follows:
\vspace{-5pt}
\begin{equation}
    \tau_{ij} = |T_{i} - T_{j}|^{-1}
\end{equation}
which is the inverse of their difference in survival duration. The more distant their survival duration is, the more SupWCon pulls their representations apart in the latent space. This is the first time in the context of survival analysis, to the best of our knowledge, that contrastive learning is used to make hardness-aware temporal distinctiveness based on the known survival duration of subjects. 

We used two more tricks that have been established as effective in the literature regarding contrastive learning. First, introducing a learnable nonlinear transformation, such as a simple two-layer fully connected neural net with a nonlinear activation function, between the ultimate instance representation and where the SupWCon loss performs. This trick substantially improves the quality of the ultimate instance representations compared to when the SupWCon performs directly on them \cite{chen2020simple}\footnote{The work in \cite{chen2020simple} conjectures that the importance of using the representation before the nonlinear projection is due to loss of information induced by the contrastive loss on the direct vectors that contrastive loss is working on.}. Second, we normalized the vector representations of instances onto the unit sphere ($l_{2}$ normalization) prior to using them in SupWCon, which also experimentally proved to be effective \cite{chen2020simple}.  

It is noteworthy that the contrastive learning component is only used during training to add hardness-aware distinctive refinements to the ultimate representations and is discarded during inference. 

\subsection{Survival Prediction}

For continuous survival models, the hazard function, denoted as $\lambda(t)$, represents the instantaneous probability of an event occurring at time $t$, given that the individual has survived up to time $t$. However, in the discrete setting, where time is considered as a sequence of distinct points, the hazard function is defined differently. Instead of dealing with infinitesimal intervals, the hazard function represents the conditional probability that the patient dies at a specific time $t$, given he/she was alive before $t$. Given that training data consists of pairs of covariates and time  $(x, t)$, our goal is to model the distribution of event times. The probability density function $f(t|x)$, the survival function $S(t|x)$, and the hazard function $ \lambda(t|x)$ respectively are defined as:
\vspace{-5pt}
\begin{equation}
    f(t|x) = P_{x}(T=t)
\end{equation}
which represents the probability mass assigned to an event occurring exactly at time $t$,

\begin{equation}
    S(t|x) = P_{x}(T>t)
\end{equation}
which gives the probability that an event has not occurred up to and including time $t$, and finally the hazard function formulation,

\begin{equation}
\begin{split}
    \lambda(t|x) &= P_{x}(t=T | T>t-1) \\
    &= \frac{f(t|x)}{S(t-1|x)} = \frac{S(t-1|x)-S(t|x)}{S(t-1|x)}
\end{split}
\end{equation}

Using the above formulation, we can rewrite the survival function formulation as follows:

\begin{equation}
    S(t|x) = (1 - \lambda(t|x))S(t-1|x)
\end{equation}

If we show the complement of hazard function by $r(t|x) = 1 - \lambda(t|x)$, we have:

\begin{equation}
\label{eq:S}
    S(t|x) = r(t|x)S(t-1|x)
\end{equation}

By recursively expanding on equation \ref{eq:S}, the survival function can be expressed as:
\begin{equation}
    S(t|x) = \prod_{s = 1}^{t}r(s|x)
\end{equation}

A feed-forward neural network (FFN) is the core of the survival prediction component, which predicts the complement of the hazard function $r(t | x)$ for all times up to $T_{max}$. Thus, the output of the survival prediction component for a patient $i$ is a vector of size $T_{max}$ as follows:
\vspace{-5pt}
\begin{equation}
    \hat{y_{i}} = [\hat{r_{i}}(t|x)]_{t=1}^{T_\text{max}}
\end{equation}

In continuous-time survival analysis, the mean lifetime of a patient or the expected value of the random variable $T$, which represents the average time until an event occurs, can be calculated by integrating the survival function over time. Mathematically, the mean lifetime $\mu$ (also known as the expected lifetime or average survival time) is derived in the following manner:

\begin{equation}
    \hat{\mu} = \int_{0}^{\infty} t \cdot f(t|x) \, dt = \int_{0}^{\infty} t \cdot (S(t|x).\lambda(t|x)) \, dt
\end{equation}

Using the technique of integration by parts, we arrive at,
    
\begin{equation}
    \hat{\mu} = \int_{0}^{\infty} S(t|x) \, dt
\end{equation}
which is the area under the survival curve. In the discrete-time formulation, we can approximate it by the sum of the survival probabilities up to $T_{max}$ as follows:

\begin{equation}
\label{eq:mu}
    \hat{\mu} \approx \sum_{t=1}^{T_\text{{max}}} \hat{S}(t|x) = \sum_{t=1}^{T_\text{{max}}} \prod_{s = 1}^{t}\hat{r}(s|x)
\end{equation}
We consider $\hat{\mu}$ as our predicted survival time duration. 

\subsection{Loss Functions}
In this section, we will expand upon different loss functions implemented to train our model. Besides the SupWCon loss which was explained in \ref{sec:cl}, we have three more losses working in combination with SupWCon. The motivation is to optimize the model with respect to the two important objectives of survival analysis: 1) accuracy in the prediction of survival duration for observed data, and 2) accurately ranking patients (both observed and censored) in terms of their risk and survival rate in different time points. 

\subsubsection{Loglikelihood Loss}
Loglikelihood loss is the main loss used to train the survival task. For observed data points, we minimize the following loss: 

\vspace{-5pt}
\begin{equation}
    L^{\text{Loglikelihood}}_{\text{ob}} = -\sum_{t=1}^{T-1}\log\hat{S}(t|x) - \sum_{t=T}^{T_\text{{max}}}\log(1-\hat{S}(t|x))
\end{equation}

and for censored data, the loss is defined as follows: 

\vspace{-5pt}
\begin{equation}
    L^{\text{Loglikelihood}}_{\text{cen}} = -\sum_{t=1}^{T}\hat{S}(t|x)
\end{equation}

\begin{equation}
    L^{\text{Loglikelihood}} = L^{\text{Loglikelihood}}_{\text{cen}} + L^{\text{Loglikelihood}}_{\text{ob}}
\end{equation}
where $T$ is either event time or censoring time. In other words, for observed data points, we maximize the summation of the survival probabilities for $1 \leq t<T$ (since the patient has survived in this time window) and minimize the summation of the survival probabilities for $ t \geq T$ (which means there is no survival starting from the occurrence of the event). For censored data points, we only maximize the summation of the survival probabilities for $1 \leq t \leq T$. In essence, for observed instances, the survival probabilities of all the time intervals are optimized, whereas for censored data, only the survival probabilities up to $T$, which is the time of censoring, are optimized. This is because, after censoring time, we do not have any information about the survival of the patients.

\subsubsection{Pairwise Ranking Loss}
We employ a pairwise ranking loss function that incorporates the concept of concordance and is based on the method used in \cite{hu2021transformer}. Such ranking losses have been widely used in the literature \cite{lee2018deephit, lee2019dynamic} for survival analysis. According to this idea, a patient who experiences an event at time $s$ should have a shorter predicted survival duration time (a higher risk) at time s compared to a patient who survives beyond time s. In other words, we want to penalize the discordant pairs. Let $T_{i}$ and $T_{j}$ represent the observed event times for patients i and j, and respectively, $T_{i} < T_{j}$. The predicted survival durations $\hat{T}_{i}$ and $\hat{T}_{j}$ (obtained from Eq. \ref{eq:mu}) are considered discordant if $\hat{T}_{i} > \hat{T}_{j}$. Our aim is to minimize the number of such discordant pairs. For every observed patient $i$ in the training set, we randomly select (with replacement) another patient $j$, ensuring that $T_{i} < T_{j}$. we only compare them with one other randomly selected data point since comparing with all the possible data points is too computationally expensive. As $T_{j}$ can be subject to censoring, the actual survival duration for patient $j$ cannot be smaller than $T_{j}$. Consequently, the difference between the predicted durations $\hat{T}_{i}$ and $\hat{T}_{j}$ should be at least $T_{j} - T_{i}$. Hence, the ranking loss formulation is as follows: 

\begin{equation}
    L^{\text{Ranking}} = \text{max}(0, (T_{j} - T_{i}) - (\hat{T}_{j} - \hat{T}_{i}))
\end{equation}

\subsubsection{Mean Squared Error (MSE) Loss}

MSE Loss penalizes wrong predicted survival duration times for only observed data points. This loss ensures that the proposed model performs well at accurate time prediction for observed patients instead of only being able to rank patients in terms of their risk. Therefore, for observed patients, MSE is calculated as follows:

\begin{equation}
    \text{MSE} = \frac{1}{N_\text{ob}} \sum_{i=1}^{N_\text{ob}}(T_i - \hat{T}_i)^2
\end{equation}
where $N_\text{ob}$ is the number of observed instances, $T_{i}$ is the true survival duration, and $\hat{T}_{i}$ is the predicted survival duration obtained as $\hat{\mu}$ from equation \ref{eq:mu}.

%% file: tex/experiments.tex
\section{Experiments}
\subsection{Dataset and Preprocessing}
We test our model on a real-world EHR dataset acquired from the University of Kansas Medical Center (KUMC) gathered from early 2009 to late 2021 for the purpose of the Acute Kidney Injury (AKI) study. In this dataset, each patient has a history of one year of hospital visits before the final hospital visit, called the onset visit, which was monitored for the occurrence of AKI. Each hospital visit comprises a collection of documented medical codes. Diagnosis codes were recorded using the International Classification of Diseases system in both the ninth and tenth Revisions (ICD-9 \& ICD-10). The prescription codes follow the RxNorm format, which provides standardized names for clinical drugs. After preprocessing, we acquired a dataset with the statistics demonstrated in table \ref{Statistical}. Since the dataset is highly imbalanced, we balance the training dataset by duplicating the observed data so we have a 50\% censored-50\% observed train set. For implementing the GRAM method, we used the hierarchical ontology of ICD-9\footnote{In preprocessing, all the ICD-10 codes were converted to ICD-9.} and the Anatomical Therapeutic Chemical (ATC) classification system respectively for diagnosis codes and prescription codes.

\begin{table}[t]
\centering
        \captionsetup{skip=3pt}
	\caption{ Dataset Statistics}
	\label{Statistical}
	\renewcommand{\arraystretch}{1.3}
	\makebox[\linewidth]{
\begin{tabular}{@{}ccccccc@{}}
\toprule

Number of patinets & 56779 \\
Number of censored patients & 47773 (84.1\%) \\
Number of observed patients & 9006 (15.8\%) \\
The average age of patients & 59.47 \\
Sex of patients distribution & (52\% M, 48\% F)\\
Average number of medical codes in a visit  & 13.29  \\
Average number of medical codes in a patient  & 66.46  \\

\hline

\end{tabular}
}
\end{table}

\subsection{Experimental Setting}
Conducting a hyperparameter tuning, we chose $128$ as the dimension of code embeddings for the ontological encoder. Two layers of transformer encoder each with two heads of multihead-attention are selected for the main encoder with a hidden dimension of $512$, which outputs a representation vector of size $256$. The survival prediction part is a three-layered fully connected neural network with hidden dimensions of $[256,128,9]$ which outputs 9 probabilities for each instance ($T_\text{max}=9$). Every probability is for a unit of time interval as one day. $T_\text{max}$ was chosen $9$ because $96\%$ of hospitalized patients were diagnosed with AKI or discharged (censored) in $9$ days. RMSprop optimizer with a learning rate of $1e-3$ and weight decay of $2e-5$ was employed for training the proposed model. For implementing SupWCon, after a thorough hyperparameter search, we chose a time window length of $2$ as the positive contrastive pairing criteria. As for the training strategy, We let the model first run for $40$ epochs with loglikelihood, ranking, and MSE losses, and then add the SupWCon loss and train for $50$ more epochs. This strategy gives us the best performance. As mentioned earlier, the contrastive component is discarded during inference. We released the GitHub implementation code of OTCSurv.\footnote{https://github.com/mohsen-nyb/OTCSurv.git}

\subsection{Evaluation Metrics}
The model was evaluated with two main metrics: the time-dependent discrimination index $C^{td}$\cite{antolini2005time}, and the Mean Absolute Error (MAE). The time-dependent discrimination index $C^{td}$, which is one of the most widely used evaluation metrics in survival analysis, is an extension of Harrell's concordance index (C-index). 
$C^{td}$, unlike the conventional c-index, assesses the model's discriminatory ability at specific time points, capturing changing predictive performance over time.
Also, we used the MAE of the predicted survival duration to express the model's performance in estimating the exact survival duration for observed data. 

\subsection{Results and Discussion}

\begin{figure}[t]
\begin{center}
\includegraphics[width=1.0\linewidth]{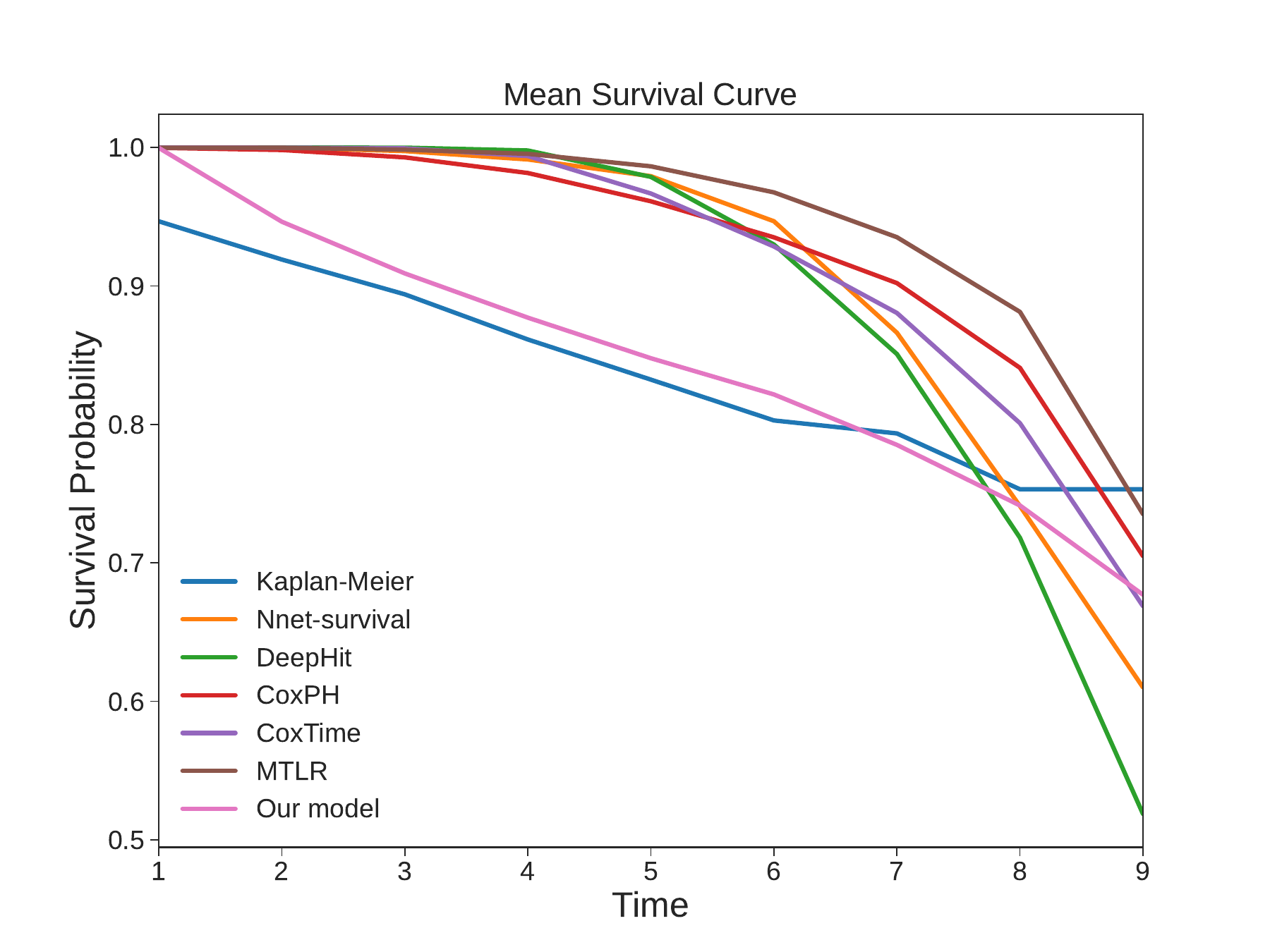}
\captionsetup{skip=1pt}
\caption{Comparison of the mean survival curves of proposed model and baselines with Kaplan-Meier curve (for all patients).}
\label{fig:mean-surv}
\end{center}
\vspace{-0.4cm}
\end{figure}

\begin{figure}[t]
\begin{center}
\includegraphics[width=1.0\linewidth]{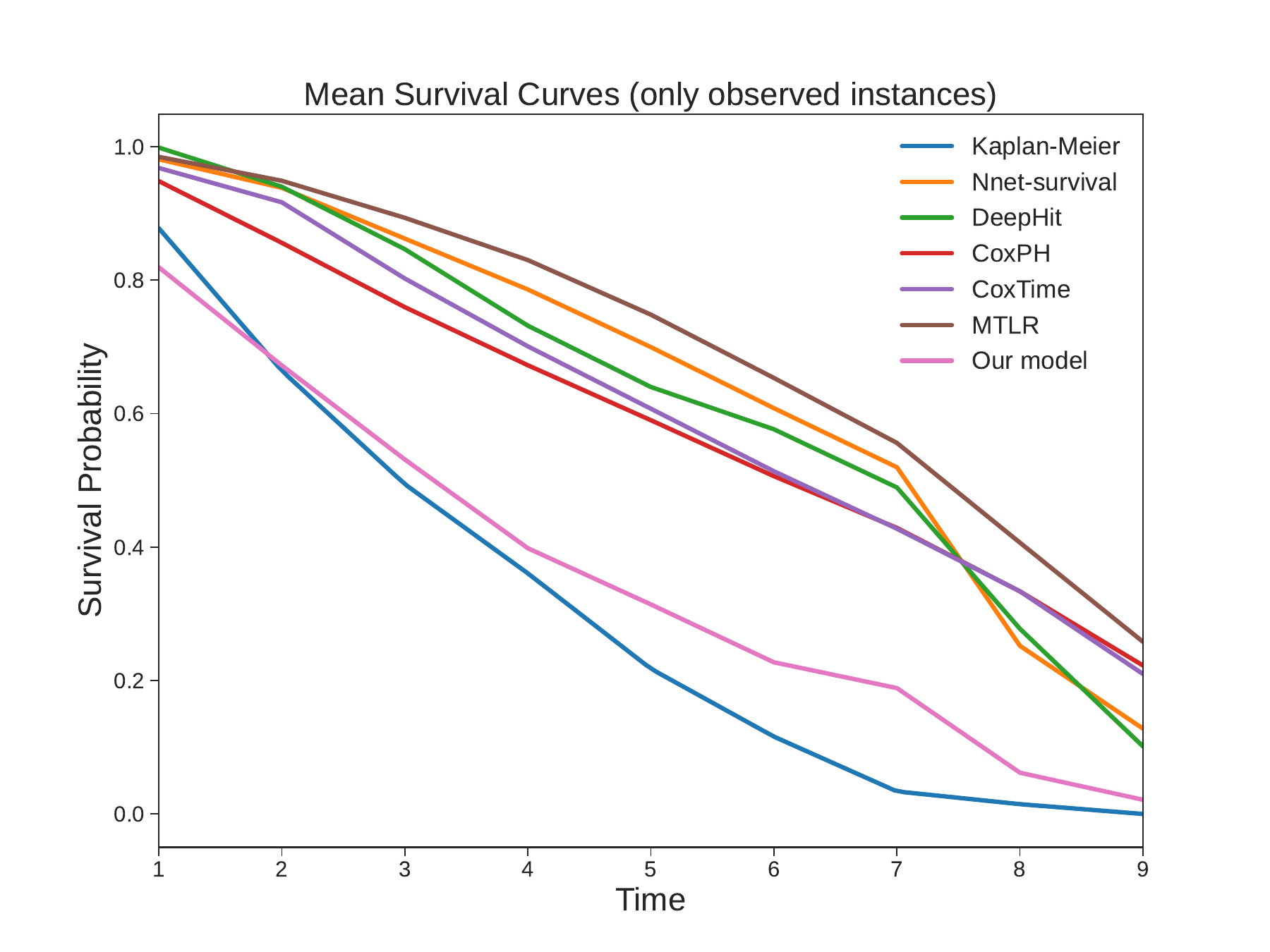}
\captionsetup{skip=1pt}
\caption{Comparison of the mean survival curves of proposed model and baselines with Kaplan-Meier curve (only for observed patients).}
\label{fig:mean-surv-ob}
\vspace{-0.4cm}
\end{center}
\end{figure}

\subsubsection{Baselines}
We compare the results of our model with various popular baselines, which are introduced below briefly. Table \ref{table2} shows the performance of each model on the AKI survival analysis task. It is evident that our proposed model outperforms all of the baselines regarding both evaluation metrics. Also, Figure \ref{fig:mean-surv} and Figure \ref{fig:mean-surv-ob} exhibit the comparison of the mean survival curves of each model with the Kaplan-Meier curve, which is the survival curve based on true data. Our proposed model's mean survival curve is the closest to the Kaplan-Meier curve considering all the data as well as only observed data, indicating that our model accurately captures the survival behavior and provides survival predictions that are more consistent with the actual outcomes.

\begin{itemize}
\item \textbf{Nnet-survival} \cite{gensheimer2019scalable}:  Nnet-survival which is trained with stochastic gradient descent employs parameterization of discrete hazards and optimization of survival likelihood and allows for non-proportional hazards.

\item \textbf{N-MTLR} \cite{fotso2018deep}:  The Neural Multi-Task Logistic Regression uses the Multi-Task Logistic Regression (MTLR) \cite{yu2011learning} model as its base and a deep learning architecture as its core.

\item \textbf{DeepHit} \cite{lee2018deephit}:  DeepHit is a deep learning-based survival analysis that uses a multi-task learning framework to simultaneously estimate the survival time and the event type probabilities, thereby handling competing risks.

\item \textbf{CoxTime} \cite{JMLR:v20:18-424}: Cox-Time is an extension of Cox regression that goes beyond the proportional hazards assumption and incorporates the concept of relative risk.

\item \textbf{DeepSurv (CoxPH)} \cite{katzman2018deepsurv}:  DeepSurv, a personalized treatment recommender system, is a Cox proportional hazards deep neural network, modeling interactions between a patient’s covariates and treatment effectiveness.

\end{itemize}

\begin{table}[ht]
\centering
        \captionsetup{skip=3pt}
	\caption{Evaluation based on $C^{td}$ and MAE for AKI survival prediction}
	\label{table2}
	\makebox[\linewidth]{
\begin{tabular}{@{}ccccccc@{}}
\toprule
Model Name                   & $C^{td}$ & MAE\\
\midrule

Nnet-survival                & 0.6332 & 3.165\\
MTLR (N-MTLR)                & 0.6712 & 4.081\\
DeepHit                      & 0.6929 & 3.012\\
CoxTime                      & 0.6912  & 2.980 \\
DeepSurv (CoxPH)                    & 0.6898 & 3.007\\
\textbf{Our model}                & \textbf{0.6990} & \textbf{1.890}\\
\hline
\end{tabular}
}
\vspace{-0.5cm}
\end{table}

\subsection{Ablation Study}
The ablation study was conducted to determine the contribution of each component in the model to the performance. We experimented with different combinations of loss components and show the results in Table \ref{table3}. Training the model with $L^{{\text{Loglikelihood}}}$ alone performs relatively poor, particularly in $C^{td}$. Having only ranking loss makes the model much stronger in terms of $C^{td}$, but the accurate time predictive ability of the model is reduced since MAE increases by $0.5$ compared to training only with $L^{{\text{Loglikelihood}}}$. Using $L^{\text{SupWCon}}$ along with $L^{{\text{Loglikelihood}}}$ increases the $C^{td}$ by $0.0161$ and decreases the MAE by $0.4$, demonstrating the prominent effectiveness of our SupWCon loss on improving both evaluation metrics. We also tried adding $L^{\text{Ranking}}$ to $L^{{\text{Loglikelihood}}}$ which results in a substantial increase in $C^{td}$ by $0.026$ but an undesired increase in MAE by $0.13$. 
The last two combinations bring the best performances. With $L^{\text{Loglikelihood}}$, $L^{\text{Ranking}}$, and $L^{\text{SupWCon}}$, we achieve the highest $C^{td}$.
With all four loss components, we achieve the best trade-off in the performance with a small compromise on $C^{td}$ but an improvement to the lowest MAE. Eventually, we utilized a weighted summation of these four losses as follows:

\begin{equation}
    L^{\text{Total}} = \lambda _{1}L^{\text{Loglikelihood}} + \lambda _{2}L^{\text{Ranking}} + \lambda _{3}L^{\text{SupWCon}} +  \lambda_{4}L^{\text{MSE}}
\end{equation}
where $\lambda _{1}$, $\lambda _{2}$, $\lambda _{3}$, $\lambda _{4}$ are hyperparameters.

From Table \ref{table5}, which is the ablation study of the ontological encoder and the attention-pooling blocks, we can realize that they play a significant role in improving the final results. We first, remove the ontological encoder from the architecture, which leads to a drop in the $C^{td}$ index and an increase in MAE, indicating the effectiveness of incorporating the knowledge domain from medical ontologies in the overall model performance. The same result happens when we remove both attention-pooling parts, which results in increasing the number of the model's parameters, making the model complex and less generalizable, and also losing the advantage of attention's performance boosting and Interpretability.

\begin{table}[t]
\centering
    \captionsetup{justification=raggedright,singlelinecheck=false} 
	\caption{Contributions of loss functions}
	\label{table3}
	\makebox[\linewidth]{
\begin{tabular}{@{}ccccccc@{}}
\toprule
Loss functions                   & $C^{td}$ & MAE\\
\midrule

$L^{\text{Loglikelihood}}$                                               & 0.6647  & 2.30 \\
$L^{\text{Ranking}}$                                                     & 0.6907 & 2.80\\
$L^{\text{Loglikelihood}} , L^{\text{SupWCon}}$                                 & 0.6808 & 1.90\\
$L^{\text{Loglikelihood}} , L^{\text{Ranking}}$                                 & 0.6951 & 2.43\\
$L^{\text{Loglikelihood}} , L^{\text{Ranking}} , L^{\text{SupWCon}}$                   & \textbf{0.7030} & \textbf{1.91}\\
$L^{\text{Loglikelihood}} , L^{\text{Ranking}} , L^{\text{SupWCon}} ,  L^{\text{MSE}}$        & \textbf{0.6990} & \textbf{1.89}\\ \hline

\end{tabular}
}
\end{table}

\begin{table}[t]
\centering
        \captionsetup{skip=3pt}
	\caption{Ontological encoder and attention-pooling contributions}
	\label{table5}
	\makebox[\linewidth]{
\begin{tabular}{@{}ccccccc@{}}
\toprule
Window size                   & $C^{td}$ & MAE\\
\midrule

w/o ontology                               & 0.6888  & 2.11 \\
w/o attention-pooling                      & 0.6795  & 2.21 \\
\textbf{Full model}                                 & \textbf{0.6990}  & \textbf{1.89} \\\hline
\end{tabular}
}
\end{table}

%% file: tex/interpretability.tex
\section{Interpretability}
Our proposed model can be interpreted by analyzing the attention weights learned in each of the model's components. In the ontological encoder, we can find the attention weights assigned to each medical code and its ancestors to realize their importance in generating the medical code embeddings. The weights learned in the visit-level attention-pooling determine the relative significance of each diagnosis or prescription code in calculating the visit-level representations. Also, by examining the attention weights learned in the instance-level attention-pooling, we can infer the relative importance of each of the visits inside the patient's medical history in composing the instance (patient) representations that will be used in the SA downstream task.

\begin{figure}[t]
\begin{center}
\includegraphics[width=1.05\linewidth]{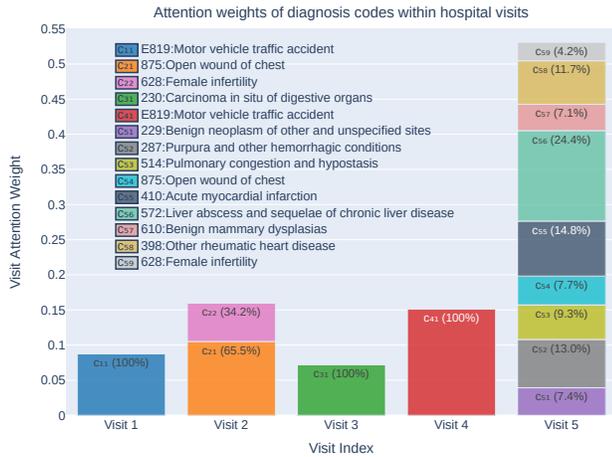}
\captionsetup{skip=1pt}
\caption{Analysis of the attention weights of the visits and the medical codes (ICD-9) inside each visit. The height of each stacked bar is the attention weight of the corresponding visit and each bar inside the stacked bars presents a medical code inside the visit and its attention weight.}
\label{fig:diag_attention}
\vspace{-0.5cm}
\end{center}
\end{figure}


To illustrate the interpretability of our model, we select a random patient diagnosed with AKI on the second day of hospitalization from the test set. Extracting the visits' attention weights from the instance-level attention-pooling and codes' attention weights from the visit-level attention-pooling, we plot Figure \ref{fig:diag_attention}. Therefore, we can realize the most important visits and the medical codes inside each visit for the model's decision-making. In Figure \ref{fig:diag_attention}, we have five stacked bars representing the five hospital visits of the patient. The height of each stacked bar shows the attention weight assigned to each visit. Each stacked bar associated with a visit has some bars indicating different codes and their attention weights. It is clear that visit 5 has the highest attention weight and consequently is the most important hospital visit for this patient. Among all the codes in this visit, "572", "410", and "287" have the highest attention weights. "572" is the ICD-9 code associated with liver abscess and sequelae of chronic liver disease, which can potentially lead to AKI \cite{duah2022acute,chancharoenthana2019acute, parikh2008long}. ICD-9 code "410" is for acute myocardial infarction (AMI), commonly known as a heart attack. AMI also can be closely associated with the onset of AKI which is discussed carefully in the medical literature \cite{shacham2019acute, wang2019risk}. ICD-9 code "287" represents purpura and other hemorrhagic conditions. Some hemorrhagic conditions, including certain types of purpura and other bleeding disorders, can potentially result in acute kidney injury (AKI) as a complication \cite{wang2017new, kelleher1987effect}. Furthermore, in Figure \ref{fig:gram}, we demonstrate how the ontological encoder learns code representations and refers to higher-level medical concepts when it comes to a rare medical code. Clearly, the "460-519" ICD-9 code which is the most general ancestor of the "514" ICD-9 code, receives the highest attention weight because first, the "514" code is not a frequent code across the train set and second, there are enough samples with the children of "460-519" (as their parent) in the train set.

%% file: tex/Conclusion.tex
\section{Conclusion}

This paper introduces a novel survival model on the basis of longitudinal healthcare data, termed Ontology-aware Temporality-based Contrastive Survival analysis (\model{}), which combines the benefits of a contrastive learning approach adapted for survival analysis as well as attention-based methods. Specifically, we designed a supervised weighted contrastive learning (SupWCon) loss function which is specifically formulated to handle data censoring and improve patients' representations using the time labels as the contrastive pairing criteria. SupWCon regulates the weights (temperature parameters) assigned to each negative pair by considering their differences in survival duration. Also, we used a sequential attention-based ontological encoder, which consists of an ontological encoder block to incorporate domain knowledge through medical ontologies, and a sequential attention encoder to capture temporal dependencies while making the model interpretable. Along with SupWCon, three other losses are employed to guide the training towards two goals of survival analysis which are risk ranking ability and precise time prediction capability. Experimental results, including baseline comparison and ablation study, on a real-world EHR dataset, showcase the superiority of the proposed model compared to existing approaches regarding both mentioned goals. Also, an attention analysis study was conducted to demonstrate the interpretability of the \model{}.

\begin{figure}[t]
\begin{center}
\includegraphics[width=0.75\linewidth]{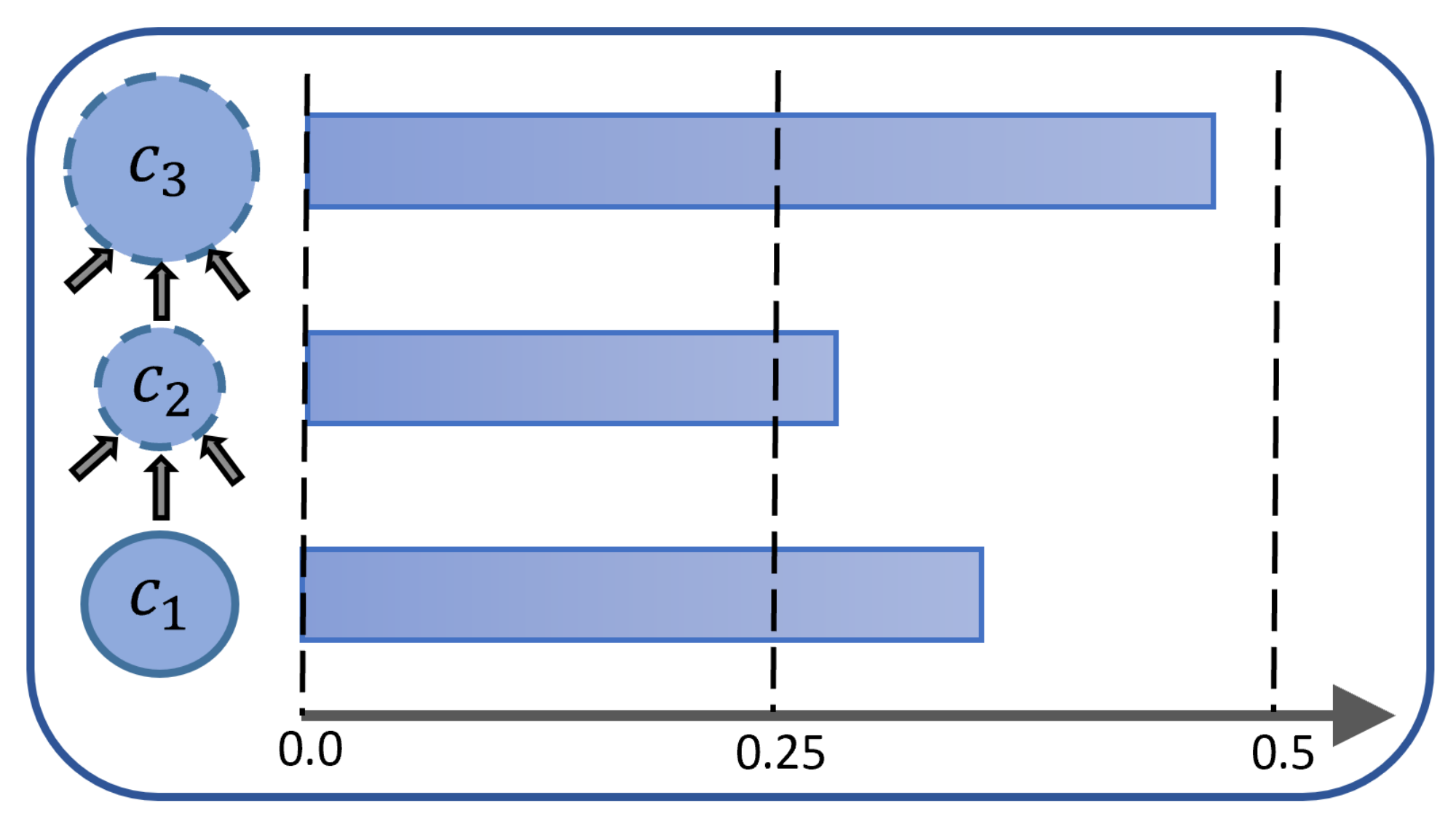}
\captionsetup{skip=5pt}
\caption{Attention weights which GRAM assigned to the ICD-9 diagnosis code $c_{1}$: 514 and its ancestors ($c_{2}$: 510-519, $c_{3}$: 460-519). The size of each node as well as the height of their bar plots show the amount of attention they received.}
\label{fig:gram}
\vspace{-0.5cm}
\end{center}
\end{figure}

\section*{Acknowledgments}
This work is partially supported by University of Kansas New Faculty Research Development (NFRD) Award, National Science Foundation Grant CNS-2125958, and WVHEPC Grant  HEPC.dsr.23.7.